\documentclass[11pt]{article}
\usepackage[margin=1in]{geometry}
\usepackage[T1]{fontenc}
\usepackage{graphicx,amsmath,amssymb,hyperref,xcolor}
\usepackage{authblk}

\begin{document}
%
%
\title{SOFA: Deep Learning Framework for Simulating and Optimizing Atrial Fibrillation Ablation}

\author[1]{Yunsung Chung}
\author[2]{Chanho Lim}
\author[2]{Ghassan Bidaoui}
\author[2]{Christian Massad}
\author[2]{Nassir Marrouche}
\author[1]{Jihun Hamm}
\affil[1]{Department of Computer Science, Tulane University, New Orleans, USA}
\affil[2]{School of Medicine, Tulane University, New Orleans, USA}
\affil[ ]{\texttt{\{ychung3,clim,gbidaoui,cmassad,nmarrouche,jhamm3\}@tulane.edu}}

\maketitle              
\begin{abstract}
Atrial fibrillation (AF) is a prevalent cardiac arrhythmia often treated with catheter ablation procedures, but procedural outcomes are highly variable. Evaluating and improving ablation efficacy is challenging due to the complex interaction between patient-specific tissue and procedural factors. This paper asks two questions: Can AF recurrence be predicted by simulating the effects of procedural parameters? How should we ablate to reduce AF recurrence? We propose SOFA (Simulating and Optimizing Atrial Fibrillation Ablation), a novel deep-learning framework that addresses these questions. SOFA first simulates the outcome of an ablation strategy by generating a post-ablation image depicting scar formation, conditioned on a patient's pre-ablation LGE-MRI and the specific procedural parameters used (e.g., ablation locations, duration, temperature, power, and force). During this simulation, it predicts AF recurrence risk. Critically, SOFA then introduces an optimization scheme that refines these procedural parameters to minimize the predicted risk. Our method leverages a multi-modal, multi-view generator that processes 2.5D representations of the atrium. Quantitative evaluations show that SOFA accurately synthesizes post-ablation images and that our optimization scheme leads to a 22.18\% reduction in the model-predicted recurrence risk. To the best of our knowledge, SOFA is the first framework to integrate the simulation of procedural effects, recurrence prediction, and parameter optimization, offering a novel tool for personalizing AF ablation. The code is available at our repository: \href{https://github.com/cys1102/SOFA}{link}.


\end{abstract}
\section{Introduction}
AF is the most common cardiac arrhythmia, which is associated with significant morbidity and mortality. Catheter ablation has emerged as a promising treatment option. However, procedural success remains highly variable due to complex patient-specific atrial anatomy and unpredictable tissue response to energy delivery. While post-ablation late gadolinium enhancement magnetic resonance imaging (LGE-MRI) is used to assess resultant scar formation, predicting the efficacy of an ablation procedure remains a major clinical challenge.

Recent advances in deep learning have demonstrated remarkable performance in medical image synthesis and predictive modeling. For example, prior studies \cite{muffoletto2021toward,muizniece2021reinforcement,ogbomo2023exploring} have used convolutional neural networks (CNNs) and reinforcement learning to simulate catheter ablation strategies, while research in AF recurrence prediction \cite{varela2017novel,atta2021new,bifulco2023explainable,razeghi2023atrial} has focused on extracting anatomical and functional features from imaging data. However, two key limitations persist. First, few methods attempt to simulate the direct effects of specific procedural choices on atrial tissue to predict outcomes. Second, selecting from predefined strategies is less nuanced than optimizing the continuous procedural parameters such as ablation duration, power, contact force, and temperature.

The importance of these parameters is underscored by clinical trial DECAAF-II~\cite{marrouche2021efficacy}, which investigated whether refining where to ablate (i.e., fibrosis-guided ablation in addition to PVI) improves outcomes. The trial's findings suggested that this location-based strategy refinement did not significantly increase success rates, highlighting \textbf{how to ablate} may be critical.

Motivated by these challenges, we ask two questions. ``Can AF recurrence be predicted by simulating the effects of procedural parameters?'' ``How should we ablate to reduce AF recurrence?'' We propose SOFA (Simulating and Optimizing Atrial Fibrillation Ablation), a deep-learning framework that directly addresses these questions. SOFA integrates a patient's pre-ablation MRI with specific procedural parameters to simulate post-ablation scar and predict recurrence. Its main novelty lies in its ability to then optimize these input parameters to minimize the predicted risk. Our contributions are:
\begin{enumerate}
    \item \textbf{Multi-modal Fusion for Post-Ablation Image Generation:} We introduce a novel fusion module that integrates pre-ablation imaging with key procedural parameters (locations, duration, temperature, power, force) to generate post-ablation images to simulate the impact of the procedure.
    \item \textbf{Simulation-Based Recurrence Prediction:} Leveraging the simulated post-ablation state, our framework predicts the risk of AF recurrence using only pre-ablation data and the intended ablation plan.
    \item \textbf{Ablation Parameter Optimization:} We propose an optimization scheme that refines ablation parameters on a patient-specific basis to minimize the predicted risk of recurrence, which offers actionable guidance for personalized treatment.
    \item \textbf{Data-Efficient 2.5D Representation:} By employing 2.5D representations, our framework achieves robust performance even with a relatively small data set of 235 patients, which results in a compact model suitable for clinical applications.
\end{enumerate}

By unifying generation, outcome prediction, and parameter optimization, SOFA provides a comprehensive decision-support tool. To our knowledge, it is the first framework to use deep learning for this purpose, directly addressing the critical question of how to ablate for improved patient outcomes.

\section{Related Work}
\subsubsection{Ablation Simulation}
Several recent studies have explored the use of deep learning for simulating catheter ablation (CA) procedures. Muffoletto et al.~\cite{muffoletto2021toward} utilize patient-specific imaging data and CNNs to make personalized predictions of CA strategies, training multiple classifiers to tailor treatment for individual AF patients. In a similar vein, Muizniece et al.~\cite{muizniece2021reinforcement} address the same problem using reinforcement learning, while Ogbomo et al.~\cite{ogbomo2023exploring} employ CNNs combined with post hoc interpretability methods to predict the success of CA simulation strategies. In contrast to these approaches that select from a set of predefined strategies, SOFA directly optimizes key procedural parameters such as duration, temperature, power, and force to tailor the ablation procedure. Moreover, our framework predicts the effects of ablation parameters by generating post-ablation images conditioned on pre-ablation images and ablation parameters, thereby providing more effective guidance for CA procedures.

\subsubsection{AF Recurrence Prediction}
Prediction of AF recurrence has been widely explored using imaging and machine learning (ML). Statistical studies~\cite{varela2017novel,atta2021new} use LA shape features for statistical prediction, while ECG-based ML methods~\cite{kwon2023machine,jiang2023artificial,zvuloni2022atrial,escribano2024combination,park2024artificial,yu2025nomogram,baskaralingam2025predicting} predict recurrence post-ablation using pre/post-procedural ECG features. Other works~\cite{bifulco2023explainable,razeghi2023atrial} leverage fibrosis patterns and post-procedural CT scans with ML classifiers, and \cite{shade2020preprocedure,roney2022predicting} combine mechanistic simulations with ML using pre-procedure imaging.  In contrast, our method leverages procedural data, including ablation parameters such as locations, duration, temperature, power, and force, combined with pre-ablation images to predict AF recurrence before ablation, which offers a novel perspective that integrates procedural context with imaging data.


\begin{figure}[!tbp]
    \centering
    \includegraphics[width=.99\linewidth]{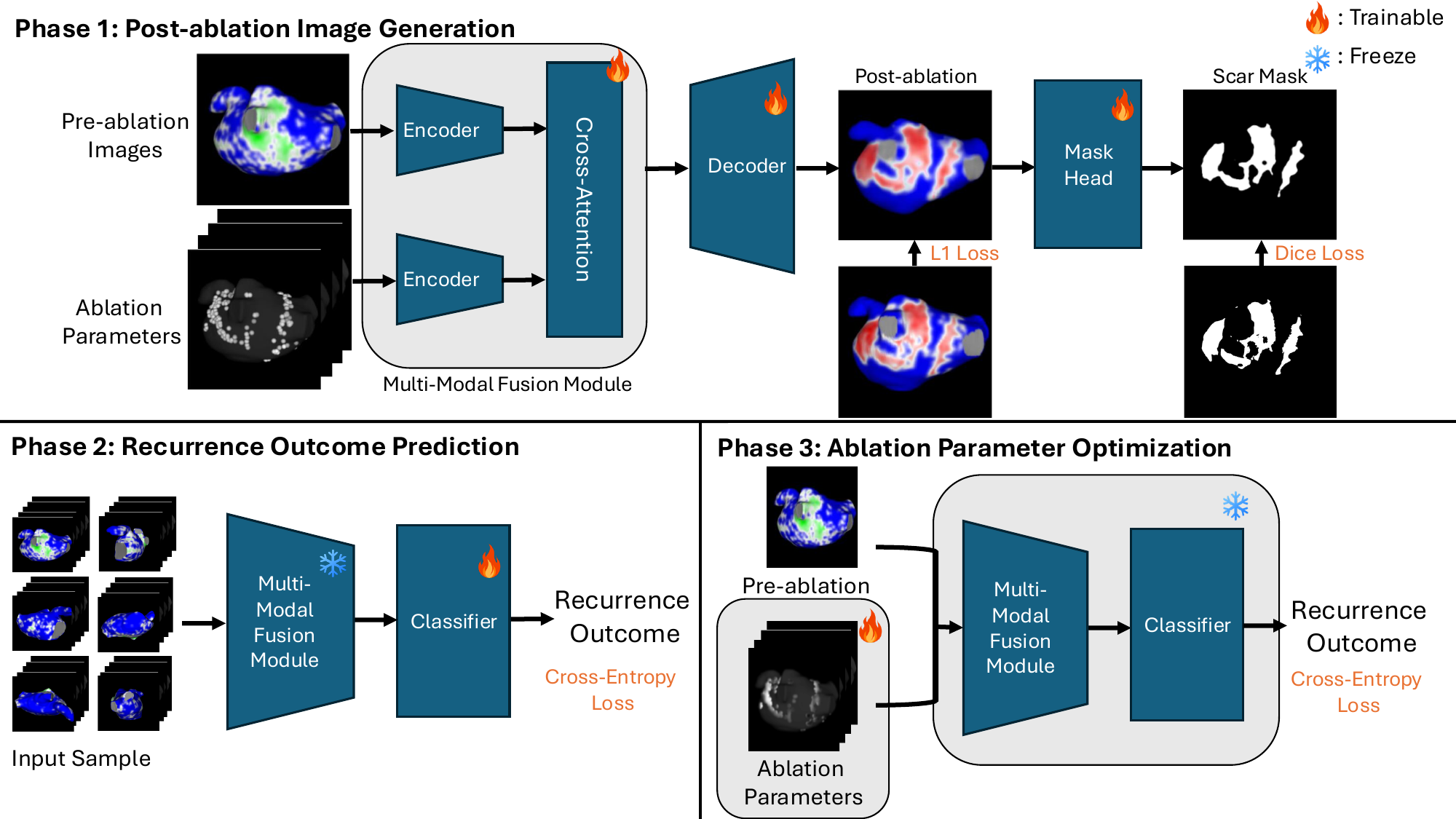}
    \caption{The overview of three phases of SOFA.
    }
    \label{fig:framework}
\end{figure}

\section{Method}

SOFA framework for simulating ablation procedures and predicting clinical outcomes consists of three sequential phases: (1) Post-ablation image generation, (2) Recurrence outcome prediction, and (3) Ablation parameter optimization. Figure~\ref{fig:framework} provides an overview of our pipeline.

\subsection{Phase 1: Image Generation and Scar Map Extraction}
Our SOFA framework utilizes two inputs, pre‐ablation RGB rendered images, \(I_{\text{pre}} \in \mathbb{R}^{3 \times H \times W}\) and ablation parameters \(I_{\text{feat}} \in \mathbb{R}^{4 \times H \times W}\). We encode these separately to produce features \(\mathbf{z}_{\text{pre}}, \mathbf{z}_{\text{feat}} \in \mathbb{R}^{C \times H_b \times W_b}\) that are then reshaped into \(\mathbf{z}_{\text{pre}}^{\text{flat}}, \mathbf{z}_{\text{feat}}^{\text{flat}} \in \mathbb{R}^{N \times C}\) (with \(N=H_bW_b\)). Our cross-attention fusion module computes queries, keys, and values using learnable weight matrices ($W_q, W_k, W_v \in \mathbb{R}^{C \times C}$):
\begin{equation}
    Q=W_q\mathbf{z}_{\text{pre}}^{\text{flat}}, \;K=W_k\mathbf{z}_{\text{feat}}^{\text{flat}}, \;V=W_v\mathbf{z}_{\text{feat}}^{\text{flat}},
\end{equation}
and fuses the features via
\begin{equation}
    \text{Attention}(Q,K,V)=\text{softmax}\Bigl(\frac{QK^\top}{\sqrt{C}}\Bigr)V,
\end{equation}
followed by a projection with \(W_o\) and then reshaping back to obtain \(B_{\text{fused}} \in \mathbb{R}^{C \times H_b \times W_b}\). 

The fused representation is processed by a decoder to produce a post-ablation image \(\hat{I}_{\text{post}} \in \mathbb{R}^{3 \times H \times W}\). A mask extraction branch then produces a scar map:
\(
\hat{M} = \sigma\Bigl( h\bigl(\hat{I}_{\text{post}}; \psi\bigr) \Bigr) \in \mathbb{R}^{1 \times H \times W},
\)
where \( h(\cdot;\psi) \) is a convolutional block and \(\sigma(\cdot)\) denotes the sigmoid activation. The overall loss function is a combination of an \( \mathcal{L}_1\) image synthesis loss and a Dice loss for scar map extraction:
\begin{equation}
    \mathcal{L}_{\text{phase1}} = \| \hat{I}_{\text{post}} - I_{\text{post}} \|_1 + \lambda\left(1 - \frac{2 \sum_{i} \hat{M}_i M_i + \epsilon}{\sum_{i} \hat{M}_i + \sum_{i} M_i + \epsilon}\right),
    \label{eq:sim_loss}
\end{equation}

where the summation index $i$ iterates over all pixels in the spatial dimensions of the masks. Here, \(\epsilon\) is a small constant for numerical stability, and \(\lambda\) is a balancing hyperparameter.

\subsection{Phase 2: AF Recurrence Outcome Prediction} 
After training the multi-modal fusion module in Phase 1, its parameters are fixed to serve as a feature extractor. For each patient, the module processes six views of the input images \(\{ I_{\text{in}}^{v} \}_{v=1}^{6}\), where \( I_{\text{in}}^{v} = [I_{\text{pre}}^{v}, I_{\text{feat}}^{v}] \), to extract a set of feature embeddings $\mathbf{z}_{v=1}^{6}$, where each $\mathbf{z}^{v}$ is generated for view $v$. These multi-view embeddings are aggregated via averaging to form a patient-level representation, which is then input to a classifier \(f(\cdot;\phi)\) for predicting the recurrence outcome $y \in \{0,1\}$ (with $1$ indicating recurrence after ablation). The prediction is given by
\begin{equation}
    \hat{y} = f\Bigl(\frac{1}{V} \{ z^{v} \}_{v=1}^{V}; \phi \Bigr).
    \label{eq:outcome_pred}
\end{equation}
where $V$ is the number of views. This classifier is trained with a binary cross-entropy loss:
\begin{equation}
    \mathcal{L}_{\text{phase2}} = - y \, \log \bigl( \sigma(\hat{y}) \bigr) - (1-y) \, \log \bigl(1 - \sigma(\hat{y})\bigr),
    \label{eq:bce_loss}
\end{equation}
where $\sigma(\cdot)$ is the sigmoid function. This approach enables outcome prediction based solely on pre-ablation data and ablation parameters, which provides early prognostic insights without requiring post-ablation imaging.

\subsection{Phase 3: Ablation Parameter Optimization}
In the final phase, we optimize the ablation parameters \( I_{\text{abl}} \in \mathbb{R}^{4 \times H \times W} \) (duration, force, temperature, and power) while keeping the pre-ablation image \( I_{\text{pre}} \in \mathbb{R}^{3 \times H \times W} \) fixed. The combined input \( I = [I_{\text{pre}}, I_{\text{abl}}] \) is processed by the fixed multi-modal fusion module to extract features that are then fed into the fixed outcome predictor \(f(\cdot;\phi)\). We optimize \( I_{\text{abl}} \) to minimize the predicted risk of recurrence by solving
\begin{equation}
    \begin{aligned}
    I_{\text{abl}}^* &= \arg \min_{I_{\text{abl}}} \, \mathcal{L}_{\text{phase3}},\\\text{where}\;\;
    \mathcal{L}_{\text{phase3}} &= -\log\Bigl( 1 - \sigma\bigl( f\bigl( [I_{\text{pre}}, I_{\text{abl}}]; \phi \bigr) \bigr) \Bigr) + \lambda_{\text{reg}} \, \| I_{\text{abl}} - I_{\text{abl}}^{0} \|_2^2.
    \end{aligned}
    \label{eq:phase3_loss}
\end{equation}
\( I_{\text{abl}}^{0} \) is the initial ablation parameter set, \( \sigma(\cdot) \) is the sigmoid function, and \( \lambda_{\text{reg}} \) (set to 0.1) balances regularization. This loss encourages \( f \) to predict a low probability of recurrence while preventing an excessive deviation from the original parameters.

To focus optimization on clinically relevant regions, we incorporate pre-generated ablation masks \( M_{\text{abl}} \in \mathbb{R}^{1 \times H \times W} \) per view, derived from the initial \( I_{\text{abl}}^{0} \) using morphological closing to smooth boundaries. These masks are binarized (\( M_{\text{abl}} > 0.5 \)) and applied during optimization to constrain updates to ablation sites. This is implemented via gradient descent
\begin{equation}
    I_{\text{abl}}^{(t+1)} = M_{\text{abl}} \cdot \Bigl(I_{\text{abl}}^{(t)} - \eta \, \nabla_{I_{\text{abl}}} \mathcal{L}_{\text{phase3}}\Bigr)
    + (1 - M_{\text{abl}}) \cdot I_{\text{abl}}^{0}.
\end{equation}
ensuring that only masked regions are modified.

\section{Experiments and Results}

\subsection{Datasets and Preprocessing}
We evaluate our framework on the DECAAF-II dataset~\cite{marrouche2021efficacy}, a randomized multicenter study designed to assess the efficacy of fibrosis-targeted ablation in patients with persistent AF. The dataset includes pre-ablation and post-ablation MRI along with detailed procedural data, including ablation points, duration, temperature, power, and force. In total, 235 patients are included, and we report 5-fold cross-validation results.

In DECAAF-II, commercial software (Merisight) was used for image segmentation, processing, quantification of left atrial fibrosis, and 3D renderings of the MRI. We preprocess these 3D models by applying rigid registration to align the left atrial models from pre- and post-ablation images, and then extract six view images per patient. All images are resized to \(256 \times 256\) pixels. Detailed implementation information is available at our anonymous repository: \href{https://anonymous.4open.science/r/SOFA-5EC0}{link}.


\begin{table}[t]
    \centering
    \caption{Phase 1 results for image simulation and scar map extraction.}
    \label{tab:phase1_results}
    \begin{tabular}{lcccc}
        \hline
        \textbf{Model} & \textbf{MSE} $\downarrow$& \textbf{PSNR} $\uparrow$& \textbf{SSIM} $\uparrow$& \textbf{Dice} $\uparrow$\\
        \hline
        Pre-ablation & 0.027 \(\pm\) 0.001 & 17.57 \(\pm\) 0.151 & 0.822 \(\pm\) 0.002 & 0.103 \(\pm\) 0.014 \\
        Ablation & 0.020 \(\pm\) 0.001 & 17.61 \(\pm\) 0.169 & 0.814 \(\pm\) 0.004 & 0.092 \(\pm\) 0.031 \\
        SOFA & \textbf{0.018 \(\pm\) 0.001} & \textbf{18.00 \(\pm\) 0.156} & \textbf{0.826 \(\pm\) 0.004} & \textbf{0.131 \(\pm\) 0.014} \\
        
        \hline
    \end{tabular}
\end{table}

\begin{table}[t]
    \centering
    \caption{Phase 2 results for recurrence prediction.}
    \label{tab:phase2_results}
    \begin{tabular}{lcc}
        \hline
        \textbf{Model} & \textbf{AUC} $\uparrow$& \textbf{Accuracy} $\uparrow$\\
        \hline
        Demographic Info & 0.578 \(\pm\) 0.051 & 0.579 \(\pm\) 0.058 \\
        Real Post-ablation & \textbf{0.711 \(\pm\) 0.039} & 0.600 \(\pm\) 0.082 \\
        SOFA & 0.671 \(\pm\) 0.052 & \textbf{0.624 \(\pm\) 0.036} \\
        \hline
    \end{tabular}
\end{table}

\begin{figure}[!tb]
    \centering
    \includegraphics[width=.90\linewidth]{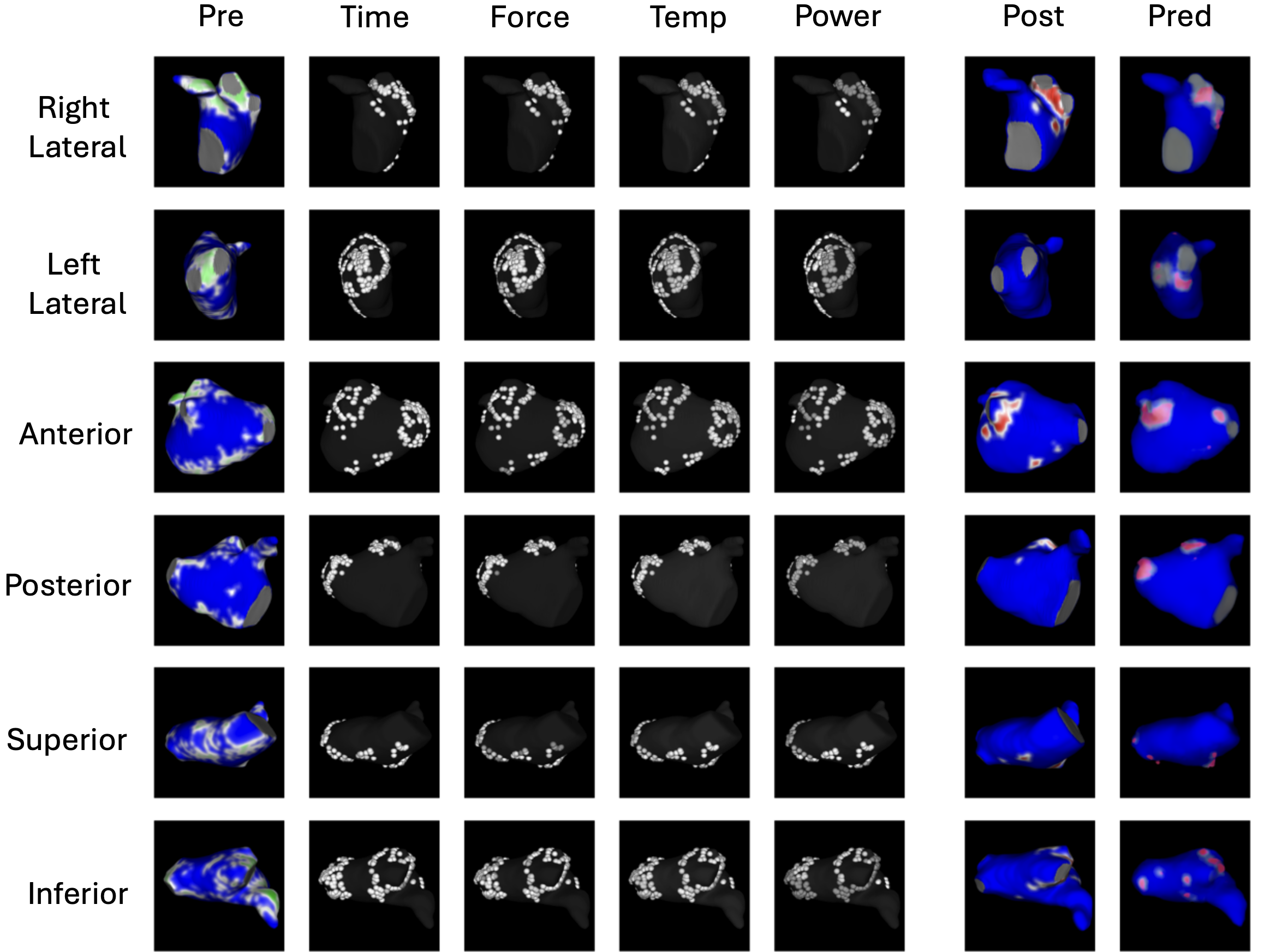}
    \caption{Phase 1 results of a patient across multiple views. Each row corresponds to a distinct view. Columns 1-5: Input data comprising pre-ablation images (Pre) and ablation parameters (Time, Force, Temp, and Power). Column 6 (Pred): Predicted post-ablation images from the generative model. Column 7 (Post): Ground-truth post-ablation images.
}
    \label{fig:phase1}
\end{figure}

\begin{figure}[!tb]
    \centering
    \includegraphics[width=.85\linewidth]{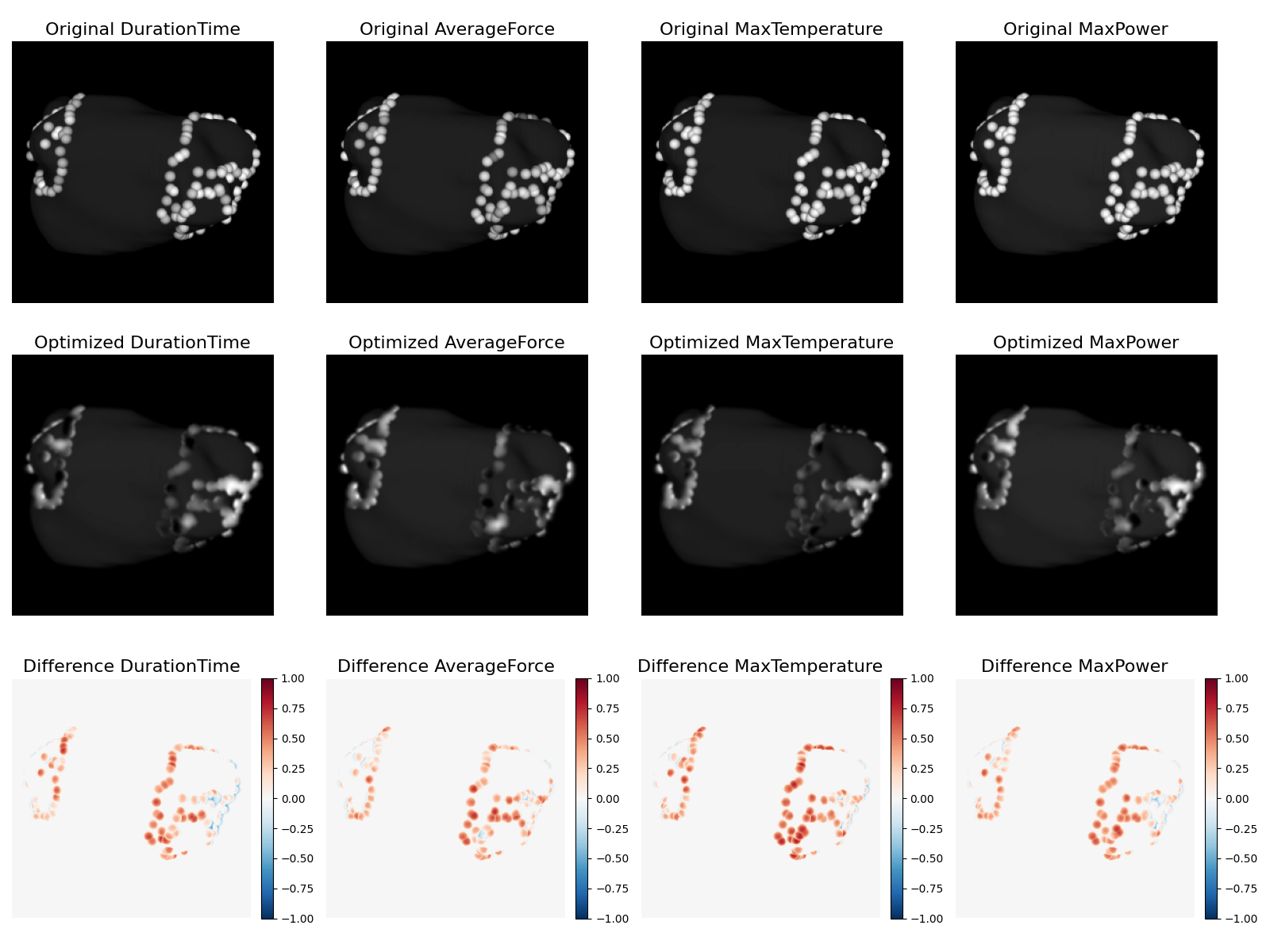}
    \caption{Phase 3 results for ablation parameter optimization. Top row: original ablation parameters. Middle row: optimized parameters. Bottom row: difference maps (red: original > optimized; blue: original < optimized).
    }
    \label{fig:phase3}
\end{figure}

\subsection{Quantitative Evaluation}
Since our task is entirely new, no prior baseline exists for direct comparison. In Tables~\ref{tab:phase1_results} and \ref{tab:phase2_results}, we present the results of SOFA alongside the baseline approaches. Table~\ref{tab:phase1_results} summarizes the performance of our image generation and scar map extraction module, where SOFA outperforms models that rely on either pre-ablation or ablation data. The improved metrics highlight the efficacy of our fusion module over traditional approaches.

In Table~\ref{tab:phase2_results}, we report the results for recurrence prediction. We compare our method with two baselines: a random forest model that uses demographic features (age, BMI, sex, etc.) and another model that employs real post-ablation images~\cite{razeghi2023atrial}. Although the post-ablation model achieves higher AUC, it is important to emphasize that our approach predicts recurrence outcomes based on pre-procedural data, which provides clinicians the opportunity to adjust treatment strategies before ablation.

Our evaluation for Phase 3 shows that the average result of recurrence was 0.487. After applying the ablation parameter optimization, the average outcome decreased to 0.379, representing a reduction of $22.18\%$ in the predicted recurrence risk. This result demonstrates the promising potential of our method to guide and optimize procedural parameters. We emphasize that while full optimization of parameters is the ideal goal, we limit ourselves to refining the given parameters rather than optimizing them entirely from scratch. Moreover, the reduction in the predicted risk is based on our model and will require future clinical validation.


\subsection{Qualitative Evaluation}
Figure~\ref{fig:phase1} illustrates the qualitative performance of our Phase 1 generative model for a single patient across six views. The model captures the overall shape and structure of the post-ablation image, which demonstrates its ability to synthesize realistic output conditioned on pre-ablation images and ablation parameters. In most views, the predicted post-ablation images (Pred) align well with the ground-truth post-ablation images (Post) with reasonable scar formations that correspond to the input parameters. For instance, in view 3, the model generates scar patterns that reflect the spatial distribution and intensity of the ablation parameters, highlighting procedural data.

However, certain limitations appear in specific views. In the right lateral, the model misses fine scar details, while the left lateral and the inferior over-generate scar tissue. Though our cross-attention fusion module excels at capturing shape and structure, it struggles with precise scar boundary delineation and parameter interplay. Future work could refine this with spatial attention in the decoder to enhance detail accuracy. 

Figure~\ref{fig:phase3} demonstrates that after optimization, the ablation parameters are adjusted in a manner that suggests longer durations and increased temperatures at some regions are associated with a lower predicted recurrence rate. The difference maps reveal that key regions exhibit an increase in these parameters. This highlights our optimization framework can potentially guide the ablation procedure to achieve improved outcomes.

\section{Conclusion}
In this work, we present SOFA, a novel multi-model, multi-view deep learning framework for simulating and optimizing AF ablation procedures. Our method integrates pre-ablation images with ablation parameters using the fusion strategy to generate post-ablation images and predict AF recurrence. Quantitative results demonstrate that our multi-modal fusion network outperforms traditional approaches in image generation and scar mask extraction, while our recurrence prediction offers valuable prognostic insights using pre-ablation data. Moreover, our ablation parameter optimization module shows promising potential by reducing the predicted recurrence rate by $22.18\%$.

A primary limitation of our study is the limited data size, which may affect the generalizability of our results. In future work, we plan to extend our framework to fully 3D data for more comprehensive spatial representations of the atrial anatomy and further enhance the clinical utility of our approach.



\bibliographystyle{splncs04}
\bibliography{miccai25}

\begin{thebibliography}{10}
\providecommand{\url}[1]{\texttt{#1}}
\providecommand{\urlprefix}{URL }
\providecommand{\doi}[1]{https://doi.org/#1}

\bibitem{atta2021new}
Atta-Fosu, T., LaBarbera, M., Ghose, S., Schoenhagen, P., Saliba, W., Tchou, P.J., Lindsay, B.D., Desai, M.Y., Kwon, D., Chung, M.K., et~al.: A new machine learning approach for predicting likelihood of recurrence following ablation for atrial fibrillation from ct. BMC Medical Imaging  \textbf{21},  1--12 (2021)

\bibitem{baskaralingam2025predicting}
Baskaralingam, A., Marchetti, M., Solana-Munoz, J., Teres, C., Le~Bloa, M., Porretta, A.P., Domenichini, G., Ascione, C., Roten, L., Knecht, S., et~al.: Predicting outcomes in persistent atrial fibrillation: the impact of surface ecg f-wave amplitude following pulmonary vein isolation. Journal of Interventional Cardiac Electrophysiology pp. 1--13 (2025)

\bibitem{bifulco2023explainable}
Bifulco, S.F., Macheret, F., Scott, G.D., Akoum, N., Boyle, P.M.: Explainable machine learning to predict anchored reentry substrate created by persistent atrial fibrillation ablation in computational models. Journal of the American Heart Association  \textbf{12}(16),  e030500 (2023)

\bibitem{escribano2024combination}
Escribano, P., R{\'o}denas, J., Garc{\'\i}a, M., Arias, M.A., Hidalgo, V.M., Calero, S., Rieta, J.J., Alcaraz, R.: Combination of frequency-and time-domain characteristics of the fibrillatory waves for enhanced prediction of persistent atrial fibrillation recurrence after catheter ablation. Heliyon  \textbf{10}(3) (2024)

\bibitem{jiang2023artificial}
Jiang, J., Deng, H., Liao, H., Fang, X., Zhan, X., Wei, W., Wu, S., Xue, Y.: An artificial intelligence-enabled ecg algorithm for predicting the risk of recurrence in patients with paroxysmal atrial fibrillation after catheter ablation. Journal of clinical medicine  \textbf{12}(5), ~1933 (2023)

\bibitem{kwon2023machine}
Kwon, S., Lee, E., Ju, H., Ahn, H.J., Lee, S.R., Choi, E.K., Suh, J., Oh, S., Rhee, W.: Machine learning prediction for the recurrence after electrical cardioversion of patients with persistent atrial fibrillation. Korean Circulation Journal  \textbf{53}(10),  677--689 (2023)

\bibitem{marrouche2021efficacy}
Marrouche, N.F., Greene, T., Dean, J.M., Kholmovski, E.G., Boer, L.M.d., Mansour, M., Calkins, H., Marchlinski, F., Wilber, D., Hindricks, G., et~al.: Efficacy of lge-mri-guided fibrosis ablation versus conventional catheter ablation of atrial fibrillation: the decaaf ii trial: study design. Journal of cardiovascular electrophysiology  \textbf{32}(4),  916--924 (2021)

\bibitem{muffoletto2021toward}
Muffoletto, M., Qureshi, A., Zeidan, A., Muizniece, L., Fu, X., Zhao, J., Roy, A., Bates, P.A., Aslanidi, O.: Toward patient-specific prediction of ablation strategies for atrial fibrillation using deep learning. Frontiers in Physiology  \textbf{12},  674106 (2021)

\bibitem{muizniece2021reinforcement}
Muizniece, L., Bertagnoli, A., Qureshi, A., Zeidan, A., Roy, A., Muffoletto, M., Aslanidi, O.: Reinforcement learning to improve image-guidance of ablation therapy for atrial fibrillation. Frontiers in Physiology  \textbf{12},  733139 (2021)

\bibitem{ogbomo2023exploring}
Ogbomo-Harmitt, S., Muffoletto, M., Zeidan, A., Qureshi, A., King, A.P., Aslanidi, O.: Exploring interpretability in deep learning prediction of successful ablation therapy for atrial fibrillation. Frontiers in Physiology  \textbf{14},  1054401 (2023)

\bibitem{park2024artificial}
Park, H., Kwon, O.S., Shim, J., Kim, D., Park, J.W., Kim, Y.G., Yu, H.T., Kim, T.H., Uhm, J.S., Choi, J.I., et~al.: Artificial intelligence estimated electrocardiographic age as a recurrence predictor after atrial fibrillation catheter ablation. NPJ Digital Medicine  \textbf{7}(1), ~234 (2024)

\bibitem{razeghi2023atrial}
Razeghi, O., Kapoor, R., Alhusseini, M.I., Fazal, M., Tang, S., Roney, C.H., Rogers, A.J., Lee, A., Wang, P.J., Clopton, P., et~al.: Atrial fibrillation ablation outcome prediction with a machine learning fusion framework incorporating cardiac computed tomography. Journal of cardiovascular electrophysiology  \textbf{34}(5),  1164--1174 (2023)

\bibitem{roney2022predicting}
Roney, C.H., Sim, I., Yu, J., Beach, M., Mehta, A., Alonso Solis-Lemus, J., Kotadia, I., Whitaker, J., Corrado, C., Razeghi, O., et~al.: Predicting atrial fibrillation recurrence by combining population data and virtual cohorts of patient-specific left atrial models. Circulation: Arrhythmia and Electrophysiology  \textbf{15}(2),  e010253 (2022)

\bibitem{shade2020preprocedure}
Shade, J.K., Ali, R.L., Basile, D., Popescu, D., Akhtar, T., Marine, J.E., Spragg, D.D., Calkins, H., Trayanova, N.A.: Preprocedure application of machine learning and mechanistic simulations predicts likelihood of paroxysmal atrial fibrillation recurrence following pulmonary vein isolation. Circulation: Arrhythmia and Electrophysiology  \textbf{13}(7),  e008213 (2020)

\bibitem{varela2017novel}
Varela, M., Bisbal, F., Zacur, E., Berruezo, A., Aslanidi, O.V., Mont, L., Lamata, P.: Novel computational analysis of left atrial anatomy improves prediction of atrial fibrillation recurrence after ablation. Frontiers in physiology  \textbf{8}, ~68 (2017)

\bibitem{yu2025nomogram}
Yu, L.j., Chen, X.H., Xu, Z., Gong, K.Z., Zhang, F.L.: A nomogram utilizing ecg p-wave parameters to predict recurrence risk following catheter ablation in paroxysmal atrial fibrillation. Journal of Cardiothoracic Surgery  \textbf{20}(1), ~94 (2025)

\bibitem{zvuloni2022atrial}
Zvuloni, E., Gendelman, S., Mohanty, S., Lewen, J., Natale, A., Behar, J.A.: Atrial fibrillation recurrence risk prediction from 12-lead ecg recorded pre-and post-ablation procedure. In: 2022 Computing in Cardiology (CinC). vol.~498, pp.~1--4. IEEE (2022)

\end{thebibliography}

\end{document}